\documentclass[runningheads]{llncs}
\usepackage{graphicx}
\usepackage{amsmath,amssymb} 
\usepackage[width=122mm,left=12mm,paperwidth=146mm,height=193mm,top=12mm,paperheight=217mm]{geometry}
\usepackage{color}
\usepackage{fixltx2e}
\usepackage{url}
\usepackage[nocompress]{cite}
\usepackage{comment}

\usepackage{ctable} 
\usepackage[pagebackref=true,breaklinks=true,letterpaper=true,bookmarks=false]{hyperref}

\newcolumntype{?}[1]{!{\vrule width #1}}

\begin{document}

\pagestyle{headings}
\mainmatter

\title{Learning to Separate Object Sounds by\\Watching Unlabeled Video}

\titlerunning{Learning to Separate Object Sounds by Watching Unlabeled Video}

\authorrunning{Ruohan Gao, Rogerio Feris, Kristen Grauman}

\author{Ruohan Gao\textsuperscript{1}, Rogerio Feris\textsuperscript{2}, Kristen Grauman\textsuperscript{3}}


\institute{\textsuperscript{1}The University of Texas at Austin, \textsuperscript{2}IBM Research, \textsuperscript{3}Facebook AI Research\\
\email{rhgao@cs.utexas.edu}, \email{rsferis@us.ibm.com}, \email{grauman@fb.com}\footnote{\emph{On leave from The University of Texas at Austin (\email{grauman@cs.utexas.edu}).}}
}

\maketitle

\begin{abstract}
Perceiving a scene most fully requires all the senses. Yet modeling how objects look and sound is challenging: most natural scenes and events contain multiple objects, and the audio track mixes all the sound sources together.  We propose to learn audio-visual object models from unlabeled video, then exploit the visual context to perform audio source separation in novel videos.  Our approach relies on a deep multi-instance multi-label learning framework to disentangle the audio frequency bases that map to individual visual objects, even without observing/hearing those objects in isolation. We show how the recovered disentangled bases can be used to guide audio source separation to obtain better-separated, object-level sounds. Our work is the first to learn audio source separation from large-scale ``in the wild" videos containing multiple audio sources per video. We obtain state-of-the-art results on visually-aided audio source separation and audio denoising. Our video results: \url{http://vision.cs.utexas.edu/projects/separating_object_sounds/}
\end{abstract}

\begin{figure}
\vspace{-0.4in}
    \center
    \includegraphics[scale=0.46]{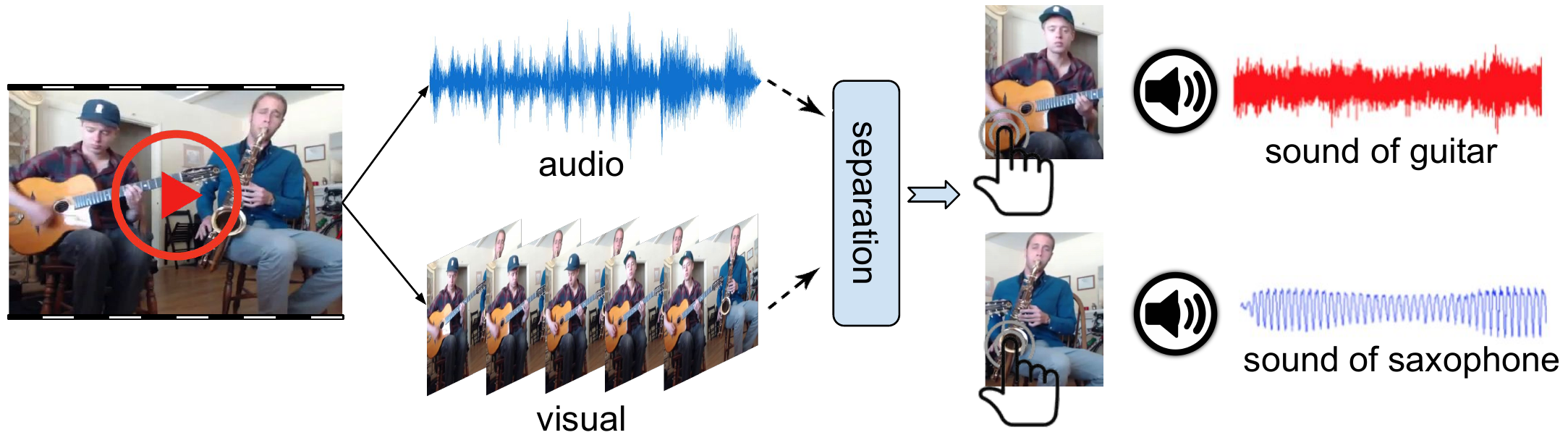}
    \label{fig:concept}
    \caption{Goal: Learn from unlabeled video to separate object sounds}
\end{figure}
\vspace{-0.15in}
\section{Introduction}
\label{sec:intro}

Understanding scenes and events is inherently a multi-modal experience.  We perceive the world by both looking and listening (and touching, smelling, and tasting).  Objects generate unique sounds due to their physical properties and interactions with other objects and the environment. For example, perception of a coffee shop scene may include seeing cups, saucers, people, and tables, but also hearing the dishes clatter, the espresso machine grind, and the barista shouting an order.  Human developmental learning is also inherently multi-modal, with young children quickly amassing a repertoire of objects and their sounds: dogs bark, cats mew, phones ring.

However, while recognition has made significant progress by ``looking"---detecting objects, actions, or people based on their appearance---it often does not listen. Despite a long history of audio-visual video indexing~\cite{smeaton2006evaluation,snoek2005multimodal,naphade2006large,jhuo2014discovering,wang2017truly}, \emph{objects} in video are often analyzed as if they were silent entities in silent environments. A key challenge is that in a realistic video, object sounds are observed not as separate entities, but as a \emph{single audio channel} that mixes all their frequencies together. Audio source separation, though studied extensively in the signal processing literature~\cite{hyvarinen2000independent,zibulevsky2001blind,virtanen2007monaural,fevotte2009nonnegative}, remains a difficult problem with natural data outside of lab settings.  Existing methods perform best by capturing the input with multiple microphones, or else assume a clean set of single source audio examples is available for supervision (e.g., a recording of only a violin, another recording containing only a drum, etc.), both of which are very limiting prerequisites. The blind audio separation task evokes challenges similar to image segmentation---and perhaps more, since all sounds overlap in the input signal.

Our goal is to learn how different objects sound by both looking at \emph{and} listening to unlabeled video containing multiple sounding objects. We propose an unsupervised approach to disentangle mixed audio into its component sound sources. The key insight is that observing sounds in a variety of visual contexts reveals the cues needed to isolate individual audio sources; the different visual contexts lend weak supervision for discovering the associations. For example, having experienced various instruments playing in various combinations before, then given a video with a guitar and a saxophone (Fig.~\ref{fig:concept}), one can naturally anticipate what sounds could be present in the accompanying audio, and therefore better separate them. Indeed, neuroscientists report that the mismatch negativity of event-related brain potentials, which is generated bilaterally within auditory cortices, is elicited only when the visual pattern promotes the segregation of the sounds~\cite{rahne2007visual}. This suggests that synchronous presentation of visual stimuli should help to resolve sound ambiguity due to multiple sources, and promote either an integrated or segregated perception of the sounds.

We introduce a novel audio-visual source separation approach that realizes this intuition. Our method first leverages a large collection of unannotated videos to discover a latent sound representation for each object.  Specifically, we use state-of-the-art image recognition tools to infer the objects present in each video clip, and we perform non-negative matrix factorization (NMF) on each video's audio channel to recover its set of frequency basis vectors.  At this point it is unknown which audio bases go with which visible object(s).  To recover the association, we construct a neural network for multi-instance multi-label learning (MIML) that maps audio bases to the distribution of detected visual objects.  From this audio basis-object association network, we extract the audio bases linked to each visual object, yielding its prototypical spectral patterns.  Finally, given a novel video, we use the learned per-object audio bases to steer audio source separation.

Prior attempts at visually-aided audio source separation tackle the problem by detecting low-level correlations between the two data streams for the input video~\cite{trevor-cocktail,fisher2001learning,barzelay2007harmony,rivet2007mixing,casanovas2010blind,parekh2017motion,pu2017audio,li2017see}, and they experiment with somewhat controlled domains of musical instruments in concert or human speakers facing the camera.  In contrast, we propose to \emph{learn object-level sound models} from hundreds of thousands of unlabeled videos, and generalize to separate new audio-visual instances. We demonstrate results for a broad set of ``in the wild" videos. While a resurgence of research on cross-modal learning from images and audio also capitalizes on synchronized audio-visual data for various tasks~\cite{kidron2005pixels,owens2016visually,owens2016ambient,aytar2016soundnet,arandjelovic2017look,arandjelovic2017objects,Korbar2018cotraining}, they treat the audio as a single monolithic input, and thus cannot associate different sounds to different objects in the same video.

The main contributions in this paper are as follows. Firstly, we propose to enhance audio source separation in videos by ``supervising" it with visual information from image recognition results\footnote{Our task can hence be seen as ``weakly supervised", though the weak ``labels" themselves are inferred from the video, not manually annotated.}. Secondly, we propose a novel deep multi-instance multi-label learning framework to learn prototypical spectral patterns of different acoustic objects, and inject the learned prior into an NMF source separation framework. Thirdly, to our knowledge, we are the first to study audio source separation learned from large scale online videos. We demonstrate state-of-the-art results on visually-aided audio source separation and audio denoising.
\vspace*{-0.2in}

\section{Related Work}
\label{sec:related}
\paragraph{Localizing sounds in video frames}
The sound localization problem entails identifying which pixels or regions in a video are responsible for the recorded sound.  Early work on localization explored correlating pixels with sounds using mutual information~\cite{hershey2000audio,fisher2001learning} or multi-modal embeddings like canonical correlation analysis~\cite{kidron2005pixels}, often with assumptions that a sounding object is in motion. Beyond identifying correlations for a single input video's audio and visual streams, recent work investigates learning associations from many such videos in order to localize sounding objects~\cite{arandjelovic2017objects}. Such methods typically assume that there is \emph{one} sound source, and the task is to localize the portion(s) of the visual content responsible for it. In contrast, our goal is to \emph{separate} multiple audio sources from a monoaural signal by leveraging learned audio-visual associations.

\vspace*{-0.1in}
\paragraph{Audio-visual representation learning}
Recent work shows that image and audio classification tasks can benefit from representation learning with both modalities. Given unlabeled training videos, the audio channel can be used as free self-supervision, allowing a convolutional network to learn features that tend to gravitate to objects and scenes, resulting in improved image classification~\cite{owens2016ambient,arandjelovic2017look}. Working in the opposite direction, the SoundNet approach uses image classifier predictions on unlabeled video frames to guide a learned audio representation for improved audio scene classification~\cite{aytar2016soundnet}. For applications in cross-modal retrieval or zero-shot classification, other methods aim to learn aligned representations across modalities, e.g., audio, text, and visual~\cite{aytar2017see}. Related to these approaches, we share the goal of learning from unlabeled video with synchronized audio and visual channels. However, whereas they aim to improve audio or image classification, our method discovers associations in order to isolate sounds per object, with the ultimate task of audio-visual source separation.

\vspace*{-0.05in}
\paragraph{Audio source separation}
Audio source separation (from purely audio input) has been studied for decades in the signal processing literature.  Some methods assume access to multiple microphones, which facilitates separation~\cite{nakadai2002real,yilmaz2004blind,duong2010under}. Others accept a single monoaural input~\cite{wang2006investigating,smaragdis2007supervised,spiertz2009source,virtanen2007monaural,huang2014deep} to perform ``blind" separation. Popular approaches include Independent Component Analysis (ICA)~\cite{hyvarinen2000independent}, sparse decomposition~\cite{zibulevsky2001blind}, Computational Auditory Scene Analysis (CASA)~\cite{ellis1996prediction}, non-negative matrix factorization (NMF)~\cite{lee2001algorithms,virtanen2007monaural,fevotte2009nonnegative,fevotte2011algorithms}, probabilistic latent variable models~\cite{hofmann1999probabilistic,smaragdis2006probabilistic}, and deep learning~\cite{huang2014deep,simpson2015deep,hershey2016deep}. NMF is a traditional method that is still widely used for unsupervised source separation~\cite{spiertz2009source,virtanen2003sound,innami2012nmf,jaiswal2011clustering,guo2015nmf}. However, existing methods typically require supervision to get good results. Strong supervision in the form of isolated recordings of individual sound sources~\cite{wang2006investigating,smaragdis2007supervised} is effective but difficult to secure for arbitrary sources in the wild. Alternatively, ``informed" audio source separation uses special-purpose auxiliary cues to guide the process, such as a music score~\cite{hennequin2011score}, text~\cite{le2015text}, or manual user guidance~\cite{bryan2014interactive,duong2014interactive,wang2006investigating}. Our approach employs an existing NMF optimization~\cite{fevotte2011algorithms}, chosen for its efficiency, but unlike any of the above we tackle audio separation informed by automatically detected visual objects.

\vspace*{-0.05in}
\paragraph{Audio-visual source separation}
The idea of guiding audio source separation using \emph{visual} information can be traced back to~\cite{fisher2001learning,trevor-cocktail}, where mutual information is used to learn the joint distribution of the visual and auditory signals, then applied to isolate human speakers.  Subsequent work explores audio-visual subspace analysis~\cite{smaragdis2003audio,pu2017audio}, NMF informed by visual motion~\cite{parekh2017motion,sedighin2016two}, statistical convolutive mixture models~\cite{rivet2007mixing}, and correlating temporal onset events~\cite{barzelay2007harmony,li2017see}.  Recent work~\cite{pu2017audio} attempts both localization and separation simultaneously; however, it assumes a moving object is present and only aims to decompose a video into background (assumed low-rank) and foreground sounds/pixels. Prior methods nearly always tackle videos of people speaking or playing musical instruments~\cite{barzelay2007harmony,rivet2007mixing,li2017see,parekh2017motion,casanovas2010blind,trevor-cocktail,fisher2001learning,pu2017audio}---domains where salient motion signals accompany audio events (e.g., a mouth or a violin bow starts moving, a guitar string suddenly accelerates).  Some studies further assume side cues from a written musical score~\cite{li2017see}, require that each sound source has a period when it alone is active~\cite{casanovas2010blind}, or use ground-truth motion captured by MoCap~\cite{parekh2017motion}.

Whereas prior work correlates low-level visual patterns---particularly motion and onset events---with the audio channel, we propose to learn from video how different \emph{objects} look and sound, whether or not an object moves with obvious correlation to the sounds.  Our method assumes access to visual detectors, but assumes no side information about a novel test video. Furthermore, whereas existing methods analyze a single input video in isolation and are largely constrained to human speakers and instruments, our approach learns a valuable prior for audio separation from a large library of \emph{unlabeled} videos. 

Concurrent with our work, other new methods for audio-visual source separation are being explored specifically for speech~\cite{owens2018audio,ephrat2018looking,afouras2018conversation,gabbay2017visual} or musical instruments~\cite{zhao2018sound}. In contrast, we study a broader set of object-level sounds including instruments, animals, and vehicles. Moreover, 
our method's training data requirements are distinctly more flexible.  
We are the first to learn from uncurated ``in the wild" videos that contain multiple objects and multiple audio sources.

\vspace*{-0.1in}
\paragraph{Generating sounds from video}
More distant from our work are methods that aim to \emph{generate} sounds from a silent visual input, using recurrent networks~\cite{owens2016visually,zhou2017visual}, conditional generative adversarial networks (C-GANs)~\cite{chen2017}, or simulators integrating physics, audio, and graphics engines~\cite{zhang2017gensound}. Unlike any of the above, our approach learns the association between how objects look and sound in order to disentangle real audio sources; our method does not aim to synthesize sounds.

\vspace*{-0.1in}
\paragraph{Weakly supervised visual learning}
Given unlabeled video, our approach learns to disentangle which sounds within a mixed audio signal go with which recognizable objects. This can be seen as a weakly supervised visual learning problem, where the ``supervision" in our case consists of automatically detected visual objects. The proposed setting of weakly supervised audio-visual learning is entirely novel, but at a high level it follows the spirit of prior work leveraging weak annotations, including early ``words and pictures" work~\cite{duygulu-translation,barnard-lda}, internet vision methods~\cite{faces-news,keywords-sudheendra}, training weakly supervised object (activity) detectors~\cite{ferrari-ijcv2012,bilen-2016,mil-cinbis,untrimmed-nets,mubarak-mil-video}, image captioning methods~\cite{karpathy-cvpr2015,donahue-cvpr2015}, or grounding acoustic units of spoken language to image regions~\cite{harwath2017learning,harwath2018jointly}. In contrast to any of these methods, our idea is to learn \emph{sound} associations for objects from unlabeled video, and to exploit those associations for audio source separation on new videos.
\vspace{-0.05 in}
\section{Approach}
\vspace{-0.05 in}

\label{sec:approach}

Our approach learns what objects sound like from a batch of unlabeled, multi-sound-source videos.  Given a new video, our method returns the separated audio channels and the visual objects responsible for them.

We first formalize the audio separation task and overview audio basis extraction with NMF (Sec.~\ref{sec:nmf}). Then we introduce our framework for learning audio-visual objects from unlabeled video (Sec.~\ref{sec:weakly}) and our accompanying deep multi-instance multi-label network (Sec.~\ref{sec:miml}).  Next we present an approach to use that network to associate audio bases with visual objects (Sec.~\ref{sec:bases}).  Finally, we pose audio source separation for novel videos in terms of a semi-supervised NMF approach (Sec.~\ref{sec:separate}).

\vspace{-0.1 in}
\subsection{Audio Basis Extraction}
\vspace{-0.05 in}

\label{sec:nmf}

Single-channel audio source separation is the problem of obtaining an estimate for each of the $J$ sources $s_j$ from the observed linear mixture $x(t)$:
	$x(t) = \sum_{j=1}^{J}s_j(t)$,
where $s_j(t)$ are time-discrete signals. The mixture signal can be transformed into a magnitude or power spectrogram $\textbf{V} \in \mathbb{R}_{+}^{F \times N}$ consisting of $F$ frequency bins and $N$ short-time Fourier transform (STFT)~\cite{griffin1984signal} frames, which encode the change of a signal's frequency and phase content over time. We operate on the frequency domain, and use the inverse short-time Fourier transform (ISTFT)~\cite{griffin1984signal} to reconstruct the sources.

Non-negative matrix factorization (NMF) is often employed~\cite{lee2001algorithms,virtanen2007monaural,fevotte2009nonnegative,fevotte2011algorithms} to approximate the (non-negative real-valued) spectrogram matrix $\textbf{V}$ as a product of two matrices $\textbf{W}$ and $\textbf{H}$:
\begin{equation}
	\textbf{V} \approx \tilde{\textbf{V}} = \textbf{W} \textbf{H},
\end{equation}
where $\textbf{W} \in \mathbb{R}_{+}^{F \times M}$ and $\textbf{H} \in \mathbb{R}_{+}^{M \times N}$.  The number of bases  $M$ is a user-defined parameter. $\textbf{W}$ can be interpreted as the non-negative audio spectral patterns, and $\textbf{H}$ can be seen as the activation matrix. Specifically, each column of $\textbf{W}$ is referred to as a \emph{basis vector}, and each row in $\textbf{H}$ represents the gain of the corresponding basis vector. The factorization is usually obtained by solving the following minimization problem:
\begin{equation}
	 \min_{\textbf{W},\textbf{H}} D(\textbf{V} | \textbf{W}\textbf{H}) \text{~subject to~} \textbf{W} \geq 0, \textbf{H} \geq 0,
\end{equation}
where $D$ is a measure of divergence, e.g., we employ the Kullback-Leibler (KL) divergence.

For each unlabeled training video, we perform NMF independently on its audio magnitude spectrogram to obtain its spectral patterns $\textbf{W}$, and throw away the activation matrix $\textbf{H}$. $M$ audio basis vectors are therefore extracted from each video.

\vspace{-0.1in}
\subsection{Weakly-Supervised Audio-Visual Object Learning Framework}

\label{sec:weakly}

Multiple objects can appear in an unlabeled video at the same time, and similarly in the associated audio track. At this point, it is unknown which of the audio bases extracted (columns of $\textbf{W}$) go with which visible object(s) in the visual frames. To discover the association, we devise a multi-instance multi-label learning (MIML) framework that matches audio bases with the detected objects.

\begin{figure}[t]
    \center
    \includegraphics[scale=0.415]{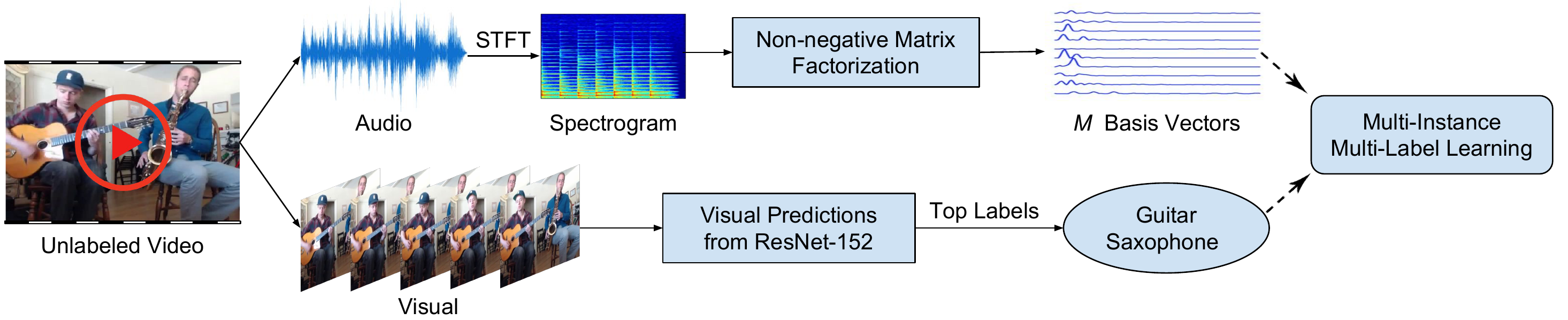}
    \vspace{-0.2in}
    \caption{Unsupervised training pipeline. For each video, we perform NMF on its audio magnitude spectrogram to get $M$ basis vectors. An ImageNet-trained ResNet-152 network is used to make visual predictions to find the potential objects present in the video. Finally, we perform multi-instance multi-label learning to disentangle which extracted audio basis vectors go with which detected visible object(s).}
    \label{fig:training_pipeline}
    \vspace{-0.05in}
\end{figure}

As shown in Fig.~\ref{fig:training_pipeline}, given an unlabeled video, we extract its visual frames and the corresponding audio track. As defined above, we perform NMF independently on the magnitude spetrogram of each audio track and obtain $M$ basis vectors from each video. For the visual frames, we use an ImageNet pre-trained ResNet-152 network~\cite{he2016deep} to make object category predictions, and we max-pool over predictions of all frames to obtain a video-level prediction. The top labels (with class probability larger than a threshold) are used as weak ``labels'' for the unlabeled video. The extracted basis vectors and the visual predictions are then fed into our MIML learning framework to discover associations, as defined next.

\vspace{-0.1in}
\subsection{Deep Multi-Instance Multi-Label Network} 
\label{sec:miml}

We cast the audio basis-object disentangling task as a multi-instance multi-label (MIML) learning problem.  In single-label MIL~\cite{dietterich1997solving}, one has bags of instances, and a bag label indicates only that some number of the instances within it have that label.  In MIML, the bag can have multiple labels, and there is ambiguity about which labels go with which instances in the bag.

\begin{figure}[t]
    \center
    \includegraphics[scale=0.5]{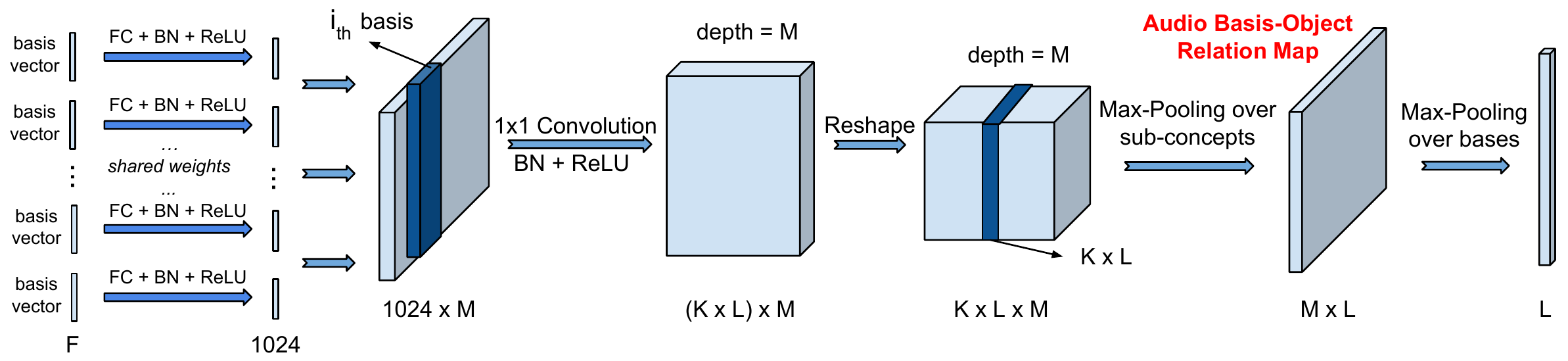}
    \vspace{-0.25in}
    \caption{Our deep multi-instance multi-label network takes a bag of $M$ audio basis vectors for each video as input, and gives a bag-level prediction of the objects present in the audio. The visual predictions from an ImageNet-trained CNN are used as weak ``labels'' to train the network with unlabeled video.}
    \label{fig:MIML_network}
     \vspace{-0.05in}
\end{figure}

We design a deep MIML network for our task. A bag of basis vectors $\{\textbf{B}\}$ is the input to the network, and within each bag there are $M$ basis vectors $\textbf{B}_i$ with $i \in [1, M]$ extracted from one video. The ``labels'' are only available at the bag level, and come from noisy visual \emph{predictions} of the ResNet-152 network trained for ImageNet recognition. The labels for each instance (basis vector) are unknown. We incorporate MIL into the deep network by modeling that there must be \emph{at least one} audio basis vector from a certain object that constitutes a positive bag, so that the network can output a correct bag-level prediction that agrees with the visual prediction.

Fig.~\ref{fig:MIML_network} shows the detailed network architecture. $M$ basis vectors are fed through a Siamese Network of $M$ branches with shared weights. The Siamese network is designed to reduce the dimension of the audio frequency bases and learns the audio spectral patterns through a fully-connected layer (FC) followed by batch norm (BN)~\cite{ioffe2015batch} and a rectified linear unit (ReLU). The output of all branches are stacked to form a $1024 \times M$ dimension feature map. Each slice of the feature map represents a basis vector with reduced dimension. Inspired by~\cite{feng2017deep}, each label is decomposed to $K$ sub-concepts to capture latent semantic meanings. For example, for drum, the latent sub-concepts could be different types of drums, such as bongo drum, tabla, and so on. The stacked output from the Siamese network is forwarded through a $1 \times 1$ Convolution-BN-ReLU module, and then reshaped into a feature cube of dimension $K \times L \times M$, where $K$ is the number of sub-concepts, $L$ is the number of object categories, and $M$ is the number of audio basis vectors. The depth of the tensor equals the number of input basis vectors, with each $K \times L$ slice corresponding to one particular basis. The activation score of the $(k,l,m)_{\text{th}}$ node in the cube represents the matching score of the $k_{\text{th}}$ sub-concept of the $l_{\text{th}}$ label for the $m_{\text{th}}$ basis vector.  

To get a bag-level prediction, we conduct two max-pooling operations. Max pooling in deep MIL~\cite{feng2017deep,wu2015deep,yang2017miml} is typically used to identify the positive instances within an aggregated bag. Our first pooling is over the sub-concept dimension ($K$) to generate an audio basis-object relation map. The second max-pooling operates over the basis dimension ($M$) to produce a video-level prediction. We use the following multi-label hinge loss to train the network:
\begin{equation}
	\mathcal{L}(A,\mathcal{V}) = \frac{1}{L}\sum_{i=1,i \neq \mathcal{V}_{j}}^{L} \sum_{j=1}^{|\mathcal{V}|} \max[0,  1 - (A_{\mathcal{V}_j} - A_i)],
 \vspace{-0.05in}
\end{equation}
where $A \in \mathbb{R}^L$ is the output of the MIML network, and represents the object predictions based on audio bases; $\mathcal{V}$ is the set of visual objects, namely the indices of the $|\mathcal{V}|$ objects predicted by the ImageNet-trained model. The loss function encourages the prediction scores of the correct classes to be larger than incorrect ones by a margin of 1. We find these pooling steps in our MIML formulation are valuable to learn accurately from the ambiguously ``labeled" bags (i.e., the videos and their object predictions); see Supp.

\vspace{-0.2in}
\subsection{Disentangling Per-Object Bases}
\vspace{-0.05in}
\label{sec:bases}
The MIML network above learns from audio-visual associations, but does not itself disentangle them. The sounds in the audio track and objects present in the visual frames of unlabeled video are diverse and noisy (see Sec.~\ref{sec:dataset} for details about the data we use). The audio basis vectors extracted from each video could be a component shared by multiple objects, a feature composed of them, or even completely unrelated to the predicted visual objects. The visual predictions from ResNet-152 network give approximate predictions about the objects that could be present, but are certainly not always reliable (see Fig.~\ref{fig:basis_label_relation} for examples).

Therefore, to collect high quality representative bases for each object category, we use our trained deep MIML network as a tool. The audio basis-object relation map after the first pooling layer of the MIML network produces matching scores across all basis vectors for all object labels. We perform a dimension-wise softmax over the basis dimension ($M$) to normalize object matching scores to probabilities along each basis dimension. By examining the normalized map, we can discover links from bases to objects. We only collect the key bases that trigger the prediction of the correct objects (namely, the visually detected objects). Further, we only collect bases from an unlabeled video if multiple basis vectors strongly activate the correct object(s). See Supp. for details, and see Fig.~\ref{fig:basis_label_relation} for examples of typical basis-object relation maps. In short, at the end of this phase, we have a set of audio bases for each visual object, discovered purely from unlabeled video and mixed single-channel audio.

\vspace{-0.1in}
\subsection{Object Sound Separation for a Novel Video}
\label{sec:separate}

\begin{figure}[t]
    \center
    \includegraphics[scale=0.555]{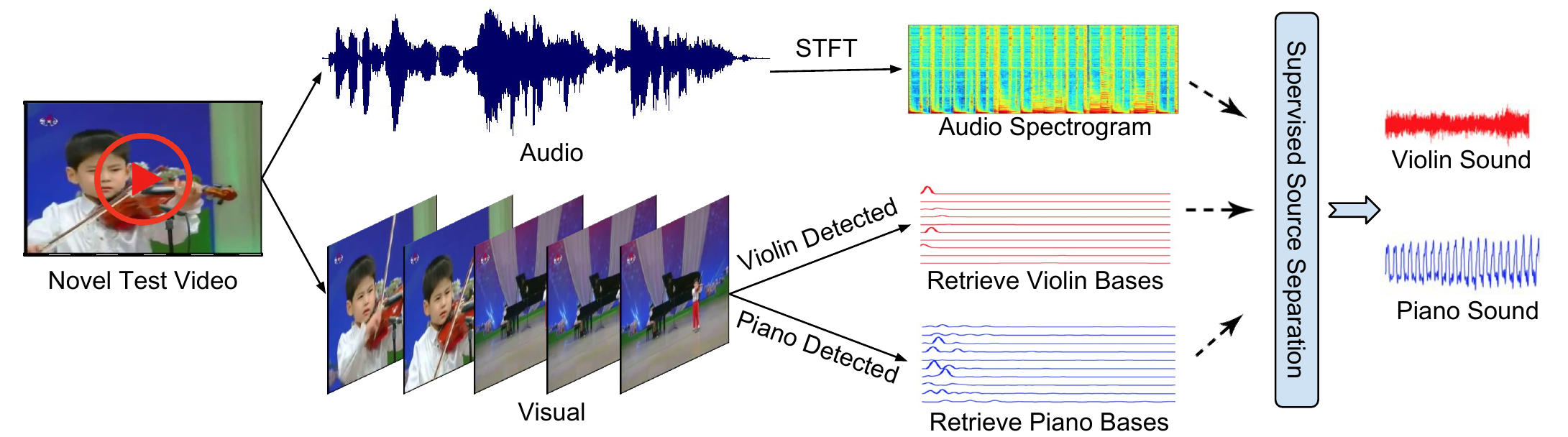}
    \vspace{-0.2in}
    \caption{Testing pipeline. Given a novel test video, we detect the objects present in the visual frames, and retrieve their learnt audio bases. The bases are collected to form a fixed basis dictionary $\textbf{W}$ with which to guide NMF factorization of the test video's audio channel. The basis vectors and the learned activation scores from NMF are finally used to separate the sound for each detected object, respectively.}
    \label{fig:testing_pipeline}
    \vspace{-0.1in}
\end{figure}

Finally, we present our procedure to separate audio sources in new videos.
As shown in Fig.~\ref{fig:testing_pipeline},  given a novel test video $q$, we obtain its audio magnitude spectrogram $\textbf{V}^{(q)}$ through STFT and detect objects using the same ImageNet-trained ResNet-152 network as before. Then, we retrieve the learnt audio basis vectors for each detected object, and use them to ``guide'' NMF-based audio source separation. Specifically, 

\vspace{-0.1in}
\begin{equation}
\label{equ:supervised_separation}
\begin{split}
		\textbf{V}^{(q)} \approx \tilde{\textbf{V}}^{(q)} & = \textbf{W}^{(q)} \textbf{H}^{(q)} \\
		& = \begin{bmatrix} \textbf{W}_1^{(q)} & \cdots & \textbf{W}_j^{(q)} & \cdots & \textbf{W}_J^{(q)}   \end{bmatrix} \begin{bmatrix} \textbf{H}_1^{(q)}  \cdots  \textbf{H}_j^{(q)}  \cdots \textbf{H}_J^{(q)}   \\ \end{bmatrix}^T,
\end{split}
\end{equation}
where $J$ is the number of detected objects ($J$ potential sound sources), and $\textbf{W}_j^{(q)}$ contains the retrieved bases corresponding to object $j$ in input video $q$. In other words, we concatenate the basis vectors learnt for each detected object to construct the  basis dictionary $\textbf{W}^{(q)}$.  Next, in the NMF algorithm, we hold $\textbf{W}^{(q)}$ fixed, and only estimate activations $\textbf{H}^{(q)}$ with multiplicative update rules.  Then we obtain the spectrogram corresponding to each detected object by $\textbf{V}_j^{(q)} = \textbf{W}_j^{(q)} \textbf{H}_j^{(q)}$.   We reconstruct the individual (compressed) audio source signals by soft masking the mixture spectrogram:
\begin{equation}
	\mathbb{V}_j = \frac{\textbf{V}_j^{(q)}}{\sum_{i=1}^J\textbf{V}_i^{(q)}} \mathbb{V},
\end{equation}
where $\mathbb{V}$ contains both magnitude and phase. Finally, we perform ISTFT on $\mathbb{V}_j$ to reconstruct the audio signals for each detected object. If a detected object does not make sound, then its estimated activation scores will be low. This phase can be seen as a self-supervised form of NMF, where the detected visual objects reveal which bases (previously discovered from unlabeled videos) are relevant to guide audio separation.
\vspace{-0.1in}
\section{Experiments}
\vspace{-0.05in}
\label{sec:exp}

We now validate our approach and compare to existing methods.

\vspace{-0.15in}
\subsection{Datasets}
\vspace{-0.05in}

\label{sec:dataset}

We consider two public video datasets: AudioSet~\cite{gemmeke2017audio} and the benchmark videos from~\cite{pu2017audio,li2014s,izadinia2013multimodal}, which we refer to as AV-Bench.

\vspace{-0.1in}
\paragraph{\textbf{AudioSet-Unlabeled:}}
We use AudioSet~\cite{gemmeke2017audio} as the source of unlabeled training videos\footnote{AudioSet offers noisy video-level audio class annotations. However, we do not use any of its label information.}. The dataset consists of short 10 second video clips that often concentrate on one event. However, our method makes no particular assumptions about using short or trimmed videos, as it learns bases in the frequency domain and pools both visual predictions and audio bases from all frames. The videos are challenging: many are of poor quality and unrelated to object sounds, such as silence, sine wave, echo, infrasound, etc. As is typical for related experimentation in the literature~\cite{arandjelovic2017objects,zhou2017visual}, we filter the dataset to those likely to display audio-visual events. In particular, we extract musical instruments, animals, and vehicles, which span a broad set of unique sound-making objects. See Supp. for a complete list of the object categories. Using the dataset's provided split, we randomly reserve some videos from the ``unbalanced'' split as validation data, and the rest as the training data. We use videos from the ``balanced'' split as test data. The final AudioSet-Unlabeled data contains 104k, 2.9k, 1k / 22k, 1.2k, 0.5k / 58k, 2.4k, 0.6k video clips in the train, val, test splits, for the
instruments, animals, and vehicles, respectively.  

\vspace{-0.1in}
\paragraph{\textbf{AudioSet-SingleSource:}}
\label{dataset:singlesource}
To facilitate quantitative evaluation (cf.~Sec.~\ref{sec:quantitative}), we construct a dataset of AudioSet videos containing only a single sounding object. We manually examine videos in the val/test set, and obtain 23 such videos. There are 15 musical instruments (accordion, acoustic guitar, banjo, cello, drum, electric guitar, flute, french horn, harmonica, harp, marimba, piano, saxophone, trombone, violin), 4 animals (cat, dog, chicken, frog), and 4 vehicles (car, train, plane, motorbike). Note that our method never uses these samples for training.
\vspace{-0.1in}
\paragraph{\textbf{AV-Bench:}} This dataset contains the benchmark videos (Violin Yanni, Wooden Horse, and Guitar Solo) used in previous studies~\cite{pu2017audio,li2014s,izadinia2013multimodal}.

\vspace{-0.15in}
\subsection{Implementation Details}
\vspace{-0.05in}

\label{sec:implementation_details}

We extract a 10 second audio clip and 10 frames (every 1s) from each video. Following common settings~\cite{arandjelovic2017look}, the audio clip is resampled at 48 kHz, and converted into a magnitude spectrogram of size $2401 \times 202$ through STFT of window length 0.1s and half window overlap. We use the NMF implementation of~\cite{fevotte2011algorithms} with KL divergence and the multiplicative update solver. We extract $M=25$ basis vectors from each audio. All video frames are resized to $256 \times 256$, and $224 \times 224$ center crops are used to make visual predictions. We use all relevant ImageNet categories and group them into 23 classes by merging the posteriors of similar categories to roughly align with the AudioSet categories; see Supp. A softmax is finally performed on the video-level object prediction scores, and classes with probability greater than 0.3 are kept as weak labels for MIML training. The deep MIML network is implemented in PyTorch with $F=2,401$, $K=4$, $L=25$, and $M=25$. We report all results with these settings and did not try other values. The network is trained using Adam~\cite{adamsolver} with weight decay $10^{-5}$ and batch size 256. The starting learning rate is set to 0.001, and decreased by 6\% every 5 epochs and trained for 300 epochs.

\vspace{-0.15in}
\subsection{Baselines}
\vspace{-0.05in}

We compare to several existing methods~\cite{spiertz2009source,lock2013joint,kidron2005pixels,pu2017audio} and multiple baselines:

\vspace*{-0.1in}
\paragraph{\textbf{MFCC Unsupervised Separation~\cite{spiertz2009source}:}}
This is an off-the-shelf unsupervised audio source separation method. The separated channels are first converted into Mel frequency cepstrum coefficients (MFCC), and then K-means clustering is used to group separated channels.
This is an established pipeline in the literature~\cite{virtanen2003sound,innami2012nmf,jaiswal2011clustering,guo2015nmf}, making it a good representative for comparison.  We use the publicly available code\footnote{\url{https://github.com/interactiveaudiolab/nussl}}.

\vspace*{-0.1in}
\paragraph{\textbf{AV-Loc~\cite{pu2017audio}, JIVE~\cite{lock2013joint}, Sparse CCA~\cite{kidron2005pixels}:}}
We refer to results reported in~\cite{pu2017audio} for the AV-Bench dataset to compare to these methods.

\vspace*{-0.1in}
\paragraph{\textbf{AudioSet Supervised Upper-Bound:}}
This baseline uses AudioSet ground-truth labels to train our deep MIML network.  AudioSet labels are organized in an ontology and each video is labeled by many categories. We use the 23 labels aligned with our subset (15 instruments, 4 animals, and 4 vehicles). This baseline serves as an upper-bound.  

\vspace*{-0.1in}
\paragraph{\textbf{K-means Clustering Unsupervised Separation:}}
We use the same number of basis vectors as our method to initialize the $\textbf{W}$ matrix, and perform unsupervised NMF. K-means clustering is then used to group separated channels, with  $K$ equal to the number of ground-truth sources. The sound sources are separated by aggregating the channel spectrograms belonging to each cluster.

\vspace*{-0.1in}
\paragraph{\textbf{Visual Exemplar for Supervised Separation:}}
We recognize objects in the frames, and retrieve bases from an exemplar video for each detected object class to supervise its NMF audio source separation. An exemplar video is the one that has the largest confidence score for a class among all unlabeled training videos.

\vspace*{-0.1in}
\paragraph{\textbf{Unmatched Bases for Supervised Separation:}}
This baseline is the same as our method except that it retrieves bases of the wrong class (at random from classes absent in the visual prediction) to guide NMF audio source separation.

\vspace*{-0.1in}
\paragraph{\textbf{Gaussian Bases for Supervised Separation:}}
We initialize the weight matrix $\textbf{W}$ randomly using a Gaussian distribution, and then perform supervised audio source separation (with $\textbf{W}$ fixed) as in Sec.~\ref{sec:separate}.

\vspace{-0.1in}
\subsection{Quantitative Results}
\label{sec:quantitative}

\begin{table}[t]
\centering
\begin{tabular}{c?{0.5mm}c|c|c|c}
                      & Instrument Pair & Animal Pair & Vehicle Pair & Cross-Domain Pair \\ \specialrule{.12em}{.1em}{.1em}
Upper-Bound &      2.05           &  0.35    &    0.60     &     2.79                 \\  \hline
K-means Clustering     &    -2.85             &   -3.76  &     -2.71     &        -3.32              \\ 
MFCC Unsupervised~\cite{spiertz2009source}       &    0.47             &  -0.21    &    -0.05     &         1.49             \\ 
Visual Exemplar          &    -2.41             &    -4.75           &   -2.21       &    -2.28        \\ 
Unmatched Bases          &    -2.12             &    -2.46           &   -1.99       &    -1.93        \\ 
Gaussian Bases          &     -8.74            &    -9.12           &    -7.39      &     -8.21       \\ 
Ours                 &       \textbf{1.83}          &    \textbf{0.23} &    \textbf{0.49}      &      \textbf{2.53}                \\ 
\end{tabular}
\vspace{0.05in}
\caption{We pairwise mix the sounds of two single source AudioSet videos and perform audio source separation. Mean Signal to Distortion Ratio (SDR in dB, higher is better) is reported to represent the overall separation performance.}
\label{Tab:synthetic}
\vspace{-0.3in}
\end{table}

\paragraph{Visually-aided audio source separation}
For ``in the wild" unlabeled videos, the ground-truth of separated audio sources never exists.
Therefore, to allow quantitative evaluation, we create a test set consisting of combined single-source videos, following~\cite{barzelay2007harmony}.  
In particular, we take pairwise video combinations from AudioSet-SingleSource~(cf.~Sec.~\ref{dataset:singlesource}) and 1) compound their audio tracks by normalizing and mixing them and 2) compound their visual channels by max-pooling their respective object predictions. Each compound video is a test video; its reserved source audio tracks are the ground truth for evaluation of separation results.

To evaluate source separation quality, we use the widely used BSS-EVAL toolbox~\cite{vincent2006performance} and report the Signal to Distortion Ratio (SDR). We perform four sets of experiments: pairwise compound two videos of musical instruments (Instrument Pair), two of animals (Animal Pair), two of vehicles (Vehicle Pair), and two cross-domain videos (Cross-Domain Pair). For unsupervised clustering separation baselines, we evaluate both possible matchings and take the best results (to the baselines' advantage).

Table~\ref{Tab:synthetic} shows the results. Our method significantly outperforms the Visual Exemplar, Unmatched, and Gaussian baselines, demonstrating the power of our learned bases. Compared with the unsupervised clustering baselines, including~\cite{spiertz2009source}, our method achieves large gains. It also has the capability to match the separated source to  acoustic objects in the video, whereas the baselines can only return ungrounded audio signals. We stress that both our method as well as the baselines use no audio-based supervision. In contrast, other state-of-the-art audio source separation methods supervise the separation process with labeled training data containing clean ground-truth sources and/or tailor separation to music/speech (e.g.,~\cite{huang2014deep,hershey2016deep,liutkus2014kernel}). Such methods are not applicable here. 

Our MIML solution is fairly tolerant to imperfect visual detection.  Using weak labels from the ImageNet pre-trained ResNet-152 network performs similarly to using the AudioSet ground-truth labels with about 30\% of the labels corrupted.  Using the true labels (Upper-Bound in Table~\ref{Tab:synthetic}) reveals the extent to which better visual models would improve results.

\vspace{-0.1in}
\paragraph{Visually-aided audio denoising}
To facilitate comparison to prior audio-visual methods (none of which report results on AudioSet), next we perform the same experiment as in~\cite{pu2017audio} on visually-assisted audio denoising on AV-Bench. Following the same setup as~\cite{pu2017audio}, the audio signals in all videos are corrupted with white noise with the signal to noise ratio set to 0 dB. To perform audio denoising, our method retrieves bases of detected object(s) and appends the same number of randomly initialized bases as the weight matrix $\textbf{W}$ to supervise NMF. The randomly initialized bases are intended to capture the noise signal. As in~\cite{pu2017audio}, we report Normalized SDR (NSDR), which measures the improvement of the SDR between the mixed noisy signal and the denoised sound.

Table~\ref{Tab:denoising} shows the results. Note that the method of~\cite{pu2017audio} is tailored to separate noise from the foreground sound by exploiting the low-rank nature of background sounds. Still, our method outperforms \cite{pu2017audio} on 2 out of the 3 videos, and performs much better than the other two prior audio-visual methods~\cite{kidron2005pixels,lock2013joint}. Pu et al.~\cite{pu2017audio} also exploit motion in manually segmented regions. On Guitar Solo, the hand's motion may strongly correlate with the sound, leading to their better performance.

\begin{table}[t]
\centering
\begin{tabular}{c?{0.5mm}c|c|c?{0.5mm}c}

           & Wooden Horse & Violin Yanni & Guitar Solo & Average \\ \specialrule{.12em}{.1em}{.1em}
Sparse CCA (Kidron et al.~\cite{kidron2005pixels}) &     4.36         &       5.30       &   5.71       &  5.12 \\ 
JIVE (Lock et al.~\cite{lock2013joint})      &       4.54       &              4.43 &   2.64       &  3.87 \\ 
Audio-Visual (Pu et al.~\cite{pu2017audio})  &     8.82         &              5.90 &   \textbf{14.1}    &     9.61  \\ \hline
Ours       &      \textbf{12.3}        &      \textbf{7.88}        &   11.4       & \textbf{10.5}  \\ 
\end{tabular}
\vspace{0.1in}
\caption{Visually-assisted audio denoising results on three benchmark videos, in terms of NSDR (in dB, higher is better).}
\label{Tab:denoising}
\vspace{-0.3in}
\end{table}

\vspace{-0.15in}
\subsection{Qualitative Results}
\vspace{-0.05in}

Next we provide qualitative results to illustrate the effectiveness of MIML training and the success of audio source separation. Here we run our method on the real multi-source videos from AudioSet.  They lack ground truth, but results can be manually inspected for quality {(see our video\textsuperscript{\ref{project_page}}).

Fig.~\ref{fig:basis_label_relation} shows example unlabeled videos and their discovered audio basis associations. For each example, we show sample video frames, ImageNet CNN visual object predictions, as well as the corresponding audio basis-object relation map predicted by our MIML network.  We also report the AudioSet audio ground truth labels, but note that they are never seen by our method. The first example (Fig.~\ref{fig:basis_label_relation}-a) has both piano and violin in the visual frames, which are correctly detected by the CNN. The audio also contains the sounds of both instruments, and our method appropriately activates bases for both the violin and piano. Fig.~\ref{fig:basis_label_relation}-b shows a man playing the violin in the visual frames, but both piano and violin are strongly activated. Listening to the audio, we can hear that an out-of-view player is indeed playing the piano. This example accentuates the advantage of learning object sounds from thousands of unlabeled videos; our method has learned the correct audio bases for piano, and ``hears'' it even though it is off-camera in this test video. Fig.~\ref{fig:basis_label_relation}-c/d show two examples with inaccurate visual predictions, and our model correctly activates the label of the object in the audio. Fig.~\ref{fig:basis_label_relation}-e/f show two more examples of an animal and a vehicle, and the results are similar. These examples suggest that our MIML network has successfully learned the prototypical spectral patterns of different sounds, and is capable of associating audio bases with object categories. 

\begin{figure}[t]
    \center
    \includegraphics[scale=0.30]{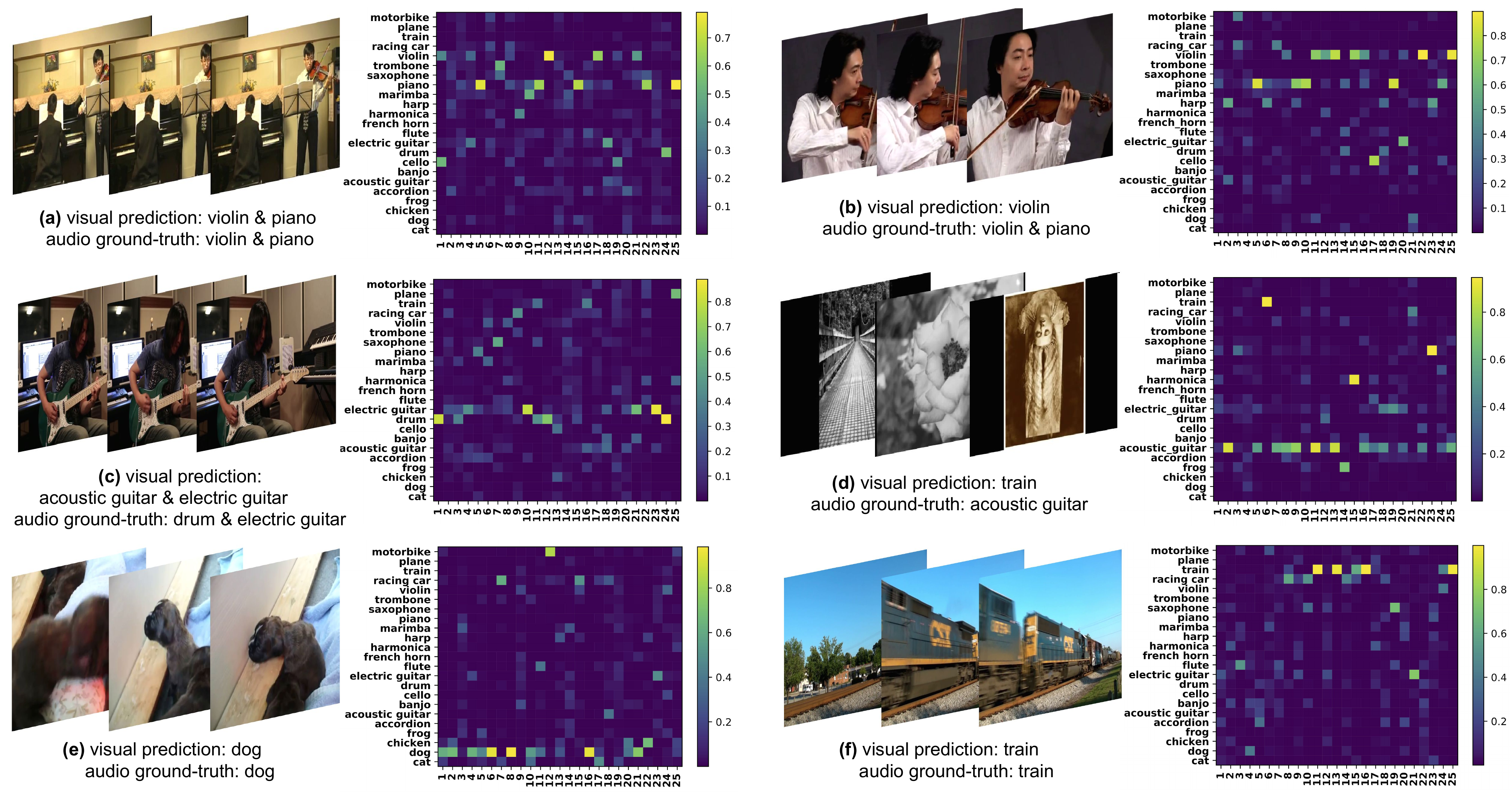}
    \vspace{-0.25in}
    \caption{In each example, we show the video frames, visual predictions, and the corresponding basis-label relation maps predicted by our MIML network. Please see our video\textsuperscript{\ref{project_page}} for more examples and the corresponding audio tracks.}
    \label{fig:basis_label_relation}
    \vspace{-0.05in}
\end{figure}

Please see our \textbf{video\footnote{\label{project_page}\url{http://vision.cs.utexas.edu/projects/separating_object_sounds/}}} for more results, where we use our system to detect and separate object sounds for novel ``in the wild" videos.

Overall, the results are promising and constitute a noticeable step towards visually guided audio source separation for more realistic videos. Of course, our system is far from perfect.  The most common failure modes by our method are when the audio characteristics of detected objects are too similar or objects are incorrectly detected (see Supp.). Though ImageNet-trained CNNs can recognize a wide array of objects, we are nonetheless constrained by its breadth. Furthermore, not all objects make sounds and not all sounds are within the camera's view. Our results above suggest that learning can be robust to such factors, yet it will be important future work to explicitly model them.
\vspace{-0.15in}
\section{Conclusion}
\vspace{-0.05in}

We presented a framework to learn object sounds from thousands of unlabeled videos. Our deep multi-instance multi-label network automatically links audio bases to object categories. Using the disentangled bases to supervise non-negative matrix factorization, our approach successfully separates object-level sounds. We demonstrate its effectiveness on diverse data and object categories. Audio source separation will continue to benefit many appealing applications, e.g., audio events indexing/remixing, audio denoising for closed captioning, or instrument equalization. In future work, we aim to explore ways to leverage scenes and ambient sounds, as well as integrate localized object detections and motion.

\noindent\textbf{Acknowledgements:}
This research was supported in part by an IBM Faculty Award, IBM Open Collaboration Research Award, and DARPA Lifelong Learning Machines. We thank members of the UT Austin vision group and Wenguang Mao, Yuzhong Wu, Dongguang You, Xingyi Zhou and Xinying Hao for helpful input. We also gratefully acknowledge a GPU donation from Facebook.

\bibliographystyle{splncs04}
\bibliography{ref_RG}
\end{document}